\def\year{2019}\relax
\documentclass[letterpaper]{article} 
\usepackage{aaai19}  
\usepackage{times}  
\usepackage{helvet} 
\usepackage{courier}  
\usepackage[hyphens]{url}  
\usepackage{graphicx} 
\usepackage{graphicx}  
\usepackage{multirow}

\title{Abstractive Summarization for Low Resource Data using Domain Transfer and Data Synthesis}
\author{Ahmed Magooda\textsuperscript{\rm 1}, Diane Litman\textsuperscript{\rm 1}\\
\textsuperscript{\rm 1} Computer Science Department, University of Pittsburgh\\ 
Pittsburgh, PA, USA\\
aem132@pitt.edu, dlitman@pitt.edu 
}
 
\begin{document}

\maketitle
\begin{abstract}
Training  abstractive summarization models typically requires large amounts of data, which can be a limitation for many domains. In this paper we explore using domain transfer and data synthesis to improve the performance of recent abstractive summarization methods when applied to small corpora of student reflections. First, we explored whether tuning state of the art model trained on newspaper data could boost performance on student reflection data. Evaluations demonstrated that summaries produced by the tuned model achieved higher ROUGE scores compared to model trained on  just student reflection data or just newspaper data. The tuned model also achieved higher scores compared to extractive summarization baselines, and additionally was judged to produce more coherent and readable  summaries in human evaluations. Second, we explored whether synthesizing summaries of student data could additionally boost performance.  We proposed a template-based  model to synthesize new data, which when incorporated into training  further increased ROUGE scores. Finally, we showed that combining data synthesis with domain transfer achieved higher ROUGE scores compared to only using one of the two approaches.
\end{abstract}

\section{Introduction}
Recently, with the emergence of neural seq2seq models, abstractive summarization methods have seen great performance strides \cite{see2017get,gehrmann2018bottom,paulus2017deep}. However, complex neural summarization models with thousands of parameters usually require a large amount of training data. In fact, much of the neural summarization work has been trained and tested in news domains where numerous large datasets exist.  For example, the CNN/DailyMail (CNN/DM) \cite{hermann2015teaching,nallapati2016abstractive} and New York Times (NYT) datasets are in the magnitude of 300k and 700k documents, respectively. In contrast, in other domains such as student reflections, summarization datasets are only in the magnitude of tens or hundreds of documents (e.g., \cite{luo2015summarizing}). We hypothesize that training complex neural abstractive summarization models in such  domains will not yield good performing models, and  we will indeed later show that this is the case for student reflections. 

To improve performance in low resource domains, we explore three directions. First, we explore domain transfer for abstractive summarization. While domain transfer is not new, compared to prior summarization studies \cite{hua2017pilot,keneshloo2019deep}, our training (news) and tuning (student reflection) domains are quite dissimilar, and the in-domain data is small. Second, we propose a template-based synthesis method to create synthesized summaries, then explore the effect of enriching training data for abstractive summarization using the proposed model compared to a synthesis baseline. Lastly, we combine both directions. Evaluations of neural abstractive summarization method across four student reflection corpora show the utility of all three methods.

\section{Related Work}

\textbf{Abstractive Summarization}. Abstractive summarization aims to generate coherent summaries with high readability, and has seen increasing interest and improved performance due to the emergence of seq2seq models \cite{sutskever2014sequence} and attention mechanisms \cite{bahdanau2014neural}. For example, \cite{see2017get}, \cite{paulus2017deep}, and \cite{gehrmann2018bottom} in addition to using encoder-decoder model with attention, they used pointer networks to solve the out of vocabulary issue, while \cite{see2017get} used coverage mechanism to solve the problem of word repetition. In addition, \cite{paulus2017deep} and \cite{P18-1063} used reinforcement learning in an end-to-end setting. 

To our knowledge, training such neural abstractive summarization models in low resource domains using domain transfer has not been thoroughly explored on domains different than news. For example, \cite{nallapati2016abstractive} reported the results of training on CNN/DM data while evaluating on DUC data without any tuning. Note that these two datasets are both in the news domain, and both consist of  well written, structured documents. 
The domain transfer experiments of \cite{gehrmann2018bottom} similarly used two different news summarization datasets (CNN/DM and NYT). 
Our work differs in several ways from these two prior domain transfer efforts. First, our experiments involve two entirely different domains: news and student reflections. Unlike news, student reflection documents lack global structure, are repetitive, and contain many sentence fragments and grammatical mistakes. Second, the prior approaches either trained a part of the model using  NYT data while retaining the other part of the model trained only on CNN/DM data \cite{gehrmann2018bottom}, or didn't perform any tuning at all  \cite{nallapati2016abstractive}. In contrast, we do the training in two consecutive phases, pretraining and fine tuning. Finally, \cite{gehrmann2018bottom} reported that while training with domain transfer outperformed training only on out-of-domain data,  it was not able to beat training only on in-domain data. This is likely because their in and out-of-domain data sizes are comparable, unlike in our case of scarce  in-domain data.

In a different approach to abstractive summarization, \cite{cao2018retrieve} developed a soft template based neural method consisting of an end-to-end deep model for template retrieval, reranking and summary rewriting. While we also develop a template based model,  our work differs in both model structure and purpose.
\\
\textbf{Data Synthesis}. Data synthesis for text summarization is underexplored, with most  prior work focusing  on machine translation, and text normalization. \cite{zhang2015character} proposed doing data augmentation through word replacement, using WordNet \cite{miller1998wordnet} and vector space similarity, respectively. We will use a WordNet replacement method as a baseline synthesis method in the experiments described below.  In contrast, \cite{fadaee2017data} synthesized/augmented data through back-translation and word replacement using language models. \cite{parida2019abstract} is another recent work that was done in parallel and is very close to ours. However, in addition to the difference in both our and their model, we think it might be infeasible to back generate student reflections from a human summary, especially an abstractive one.

\section{Reflection Summarization Dataset} 
Student reflections are comments provided by students in response to a set of instructor prompts. The prompts are directed towards gathering students' feedback on course material. 
Student reflections are collected directly following each of a set of classroom lectures over  a semester. In this paper, the set of reflections for each prompt in each lecture is considered a {\it student reflection document}. The objective of our work is to provide a comprehensive and meaningful abstractive summary of each student reflection document. Our dataset consists of documents and  summaries from four course instantiations: ENGR\footnote{http://www.coursemirror.com/download/dataset} (Introduction to Materials Science and Engineering), Stat2015 and Stat2016\footnote{http://www.coursemirror.com/download/dataset2} (Statistics for Industrial Engineers, taught in 2015 and 2016, respectively),  and CS\footnote{This data was collected and summarized by us, following the  procedures published for the  downloadable data.} (Data Structures in Computer Science). All reflections were collected in response to two pedagogically-motivated prompts~\cite{menekse2011effectiveness}:
``Point of Interest (POI): 
Describe what you found most interesting in today's class'' and 
``Muddiest Point (MP): 
Describe what was confusing or needed more detail.''

\begin{table}[t]
\begin{center}
\small
\begin{tabular}{|p{0.9\linewidth}|}
\hline \textbf{Prompt}\\
\hline Point of Interest (POI): Describe what you found most interesting in today's class.\\
\hline \textbf{Student Reflection Document}\\
\hline
\textbullet Learning about bags was very interesting.\\
\textbullet Bags as a data type and how flexible they are.\\
etc...\\
\hline \textbf{Reference Summary}\\\hline
Students were interested in ADT Bag, and also its array implementation. Many recognized that it should be resizable, and that the underlying array organization should support that. Others saw that order does not matter in bags. Some thought methods that the bag provides were interesting.\\
\hline
\end{tabular}
\end{center}
\caption{\label{tab:summaries_example} Sample data from the CS course.}
\end{table} 

\begin{table}[t]
\begin{center}
\small
\begin{tabular}{|c|c|c|c|c|}
\hline 
& \textbf{CS} & \textbf{ENGR} & \textbf{Stat2015} & \textbf{Stat2016} \\\hline
Lectures & 23 & 26 & 22 & 23\\
Prompts& 2 & 2 & 2 & 2\\
Reflections & 26 & 66 & 41 & 44\\
Summaries & 3 & 1 & 2 & 2\\
\hline
Documents & 138 & 52 & 88 & 92 \\
\hline
\end{tabular}
\end{center}
\caption{\label{dataset_summary} Dataset summary (n=370 documents).}
\end{table}

For each reflection document, at least one human (either a TA or domain expert) created summaries. 
Table \ref{tab:summaries_example} shows example reference summary produced by one annotator for the CS course. 
Table \ref{dataset_summary} summarizes the dataset in terms of number of lectures, number of prompts per lecture, average number of reflections per prompt, and number of abstractive reference summaries for each set of reflections. 

\section{Explored Approaches for Limited Resources }
To overcome the size issue of the student reflection dataset, we first explore the effect of incorporating {\bf domain transfer} into a recent abstractive summarization model: pointer networks with coverage mechanism  (PG-net)\cite{see2017get}\footnote{We also performed experiments using another recent model, fast abstractive summarization with reinforcement learning (Fast-RL)\cite{P18-1063}. Fast-RL showed similar behavior to PG-net with lower performance. Thus, due to page limit, we only report PG-net experiments}.  
To experiment with domain transfer, the model was  pretrained using the CNN/DM dataset, then fine tuned using the student reflection dataset (see the Experiments section). 
A second approach we explore to overcome the lack of reflection data is {\bf data synthesis}. 
We first propose a template model for synthesizing new data, then investigate the performance impact of using this data when training the summarization model. The proposed model makes use of the nature of datasets such as ours, where the reference summaries tend to be close in structure: humans try to find the major points that students raise, then present the points in a way that marks their relative importance (recall the CS example  in Table~\ref{tab:summaries_example}).
Our third explored approach is to {\bf combine domain transfer with data synthesis}.

\section{Proposed Template-Based Synthesis Model} 
Our motivation for using templates for data synthesis is that seq2seq synthesis models (as discussed in related work) tend to generate irrelevant and repeated words \cite{koehn2017six},
while templates can produce more coherent and concise output. 
Also, extracting templates can be done either manually or automatically typically by training a few parameters  or even doing no training, 
then external information in the form of keywords or snippets can be  populated into the templates with the help of more sophisticated models. Accordingly, using templates can be very tempting for domains with limited resources such as ours.

{\bf Model Structure.} The model consists of 4 modules:\\
\textit{1. Template extraction}: To convert human summaries into templates, we remove keywords in the summary to leave only non-keywords. We use  Rapid Automatic Keyword Extraction  (RAKE) \cite{rose2010automatic} to identify keywords.\\
\textit{2. Template clustering}: Upon converting human summaries into templates, we cluster them into $N$ clusters with the goal of using any template from  the same cluster interchangeably. 
A template is first converted into embeddings using a pretrained BERT model\footnote{https://github.com/google-research/bert\#pre-trained-models} \cite{devlin2018bert}, where template embedding is constructed by average pooling word embeddings. Templates are then clustered using k-medoid.\\
\textit{3. Summary rewriting}: An encoder-attention-decoder with pointer network is trained to perform the rewriting task. The model is trained to inject keywords into a template and perform rewriting into a coherent paragraph. The produced rewrites are considered as candidate summaries.\\
\textit{4. Summary selection}: After producing candidate summaries, we need to pick the best ones. We argue that the best candidates are those that are coherent and also convey the same meaning as the original human summary. 
We thus use a hybrid metric to score candidates, where the metric is a weighted sum of two scores and is calculated using Equations 1, 2, and 3.  Eq.1  measures coherency using a language model (LM),  Eq.2 measures how close a candidate is to a human summary using ROUGE scores, while Eq.3 picks the highest scored $N$ candidates as the final synthetic set.

\begin{equation}
\small
LM_S = (\sum_{w\in CS}log(P(w)))/(len(CS))
\end{equation}

\begin{equation}
\small
R_S = Avg(\sum_{i\in [1, 2, l]}R_{i}(CS,HS))
\end{equation}

\begin{equation}
\small
Score =(\alpha LM_S + \beta R_S)/(\alpha + \beta)
\end{equation}
CS and HS are a candidate  and human summary. $P(w)$ is the probability of word $w$ using a language model. $\alpha, \beta $ are weighting parameters. In this work we use $\alpha=\beta=1$ for all experiments. $R_{i}(CS,HS)$ is ROUGE-i score between CS and HS for i=1, 2, and $l$.

{\bf Model Training.} Before using the synthesis model, some of the constructing modules (rewriting module,  scoring LM) need training. To train the rewriting model, we use another dataset consisting of a set of samples, where each sample can be a text snippet  (sentence, paragraph, etc.). For each sample, keywords are extracted using RAKE, then removed. The keywords plus the sample with no keywords are then passed to the rewriting model. The training objective of this model is to reconstruct the original sample, which can be seen as trying to inject extracted keywords back into a template.

{\bf Model Usage.} To use the synthesis model to generate new samples, the set of human summaries are fed to the model, passing through the sub-modules in the following order:\\
1. Human summaries first pass through the template extraction module, converting each summary $s_i$ into template $t_i$ and the corresponding keywords $kw_i$.\\
2. Templates are then passed to the clustering module, producing a set of clusters. Each cluster $C$ contains a number of similar templates.\\
3. For each template $t_i$ and corresponding keywords $kw_i$ from step 1, find the cluster $C_i$ that contains the template $t_i$, then pass the set of templates within that clusters $\{t_j\} \forall{j},$ if $t_j \in C_i$ alongside the keywords $kw_i$ to the summary rewriting module. This will produce a set of candidate summaries.\\
4. The summary selection module scores and selects the highest $N$ candidates as the synthetic summaries.

\section{Experiments} \label{sec:experiments}
Our experimental designs 
 address the following hypotheses:\\
    \textbf{Hypothesis 1 (H1)} : Training complex abstractive models with limited in-domain or large quantities of out-of-domain data won't be enough to outperform extractive baselines.\\
    \textbf{Hypothesis 2 (H2)} : Domain transfer helps abstractive models even if in-domain and out-of-domain data are very different and the amount of in-domain data is very small.\\
    \textbf{Hypothesis 3 (H3)} : Enriching abstractive training data with synthetic data helps overcome in-domain data scarcity.\\
    \textbf{Hypothesis 4 (H4)} : The proposed template-based synthesis model  outperforms a simple word replacement model.\\
    \textbf{Hypothesis 5 (H5)} : Combining domain transfer with data synthesis  outperforms using each approach on its own.\\
    \textbf{Hypothesis 6 (H6)} : The synthesis model can be extended to perform reflection summarization directly. 
\\\\
\textbf{Extractive Baselines (for testing H1)}.
While \cite{see2017get}  used Lead-3 as an extractive baseline, in our data  sentence order doesn't matter as reflections are independent. We thus use a similar in concept  baseline: randomly select N reflections. Since the baseline is random we report the average result of 100 runs. Following \cite{luo2015summarizing}, we compare results to MEAD \cite{radev2004centroid} and to \cite{luo2015summarizing}'s extractive phrase-based model. Since these models extracted 5 phrases as extractive summary, we use N=5 for our three extractive baselines. Additionally we compare to running only the extractive part of Fast-RL.
\\
\textbf{Domain Transfer (for testing H2, H5)}.
\begin{table}[t]
\begin{center}
\small
\begin{tabular}{|p{0.17\linewidth}||p{0.7 \linewidth}|}
\hline \textbf{Model} & \textbf{Summary}\\
\hline  CNN/DM & Internal vs. external version of iteration iterarors i was a bit preoccupied today but seeing merge sort. How typically iterating through a linked list can be very inefficient the implementation of iterators iterators and their effectiveness how iterators can be used.\\
\hline Student Reflections & Most students found the data of data along with its mean and effectiveness interesting, as well as topics related to sse, their, and different. Students also found different a good topic.\\
\hline  Tuned & Most of students were interested in iterators, the concept of iterators, and quick sort and merge sort. They also found analyzing linked lists in regards to runtime to be interesting.\\
\hline
\multicolumn{2}{|c|}{\textbf{Human Reference}}\\
\hline
\multicolumn{2}{|p{0.9\linewidth}|}{Most of the students found iterators and linked lists as interesting. Some of them liked merge sort and quick sort. A few of them liked internal vs external iteration and analyzing runtimes of linked lists.}\\
\hline
\multicolumn{2}{|c|}{\textbf{Extracted Keywords}}\\
\hline
\multicolumn{2}{|p{0.9\linewidth}|}{iterators, linked lists, merge sort, quick sort, external iteration, analyzing runtimes}\\
\hline
\multicolumn{2}{|c|}{\textbf{Synthesized Sample}}\\\hline
\multicolumn{2}{|p{0.9\linewidth}|}{In this lecture students were mainly interested by iterators and linked lists and merge sort . they also liked quick sort, external iteration , and analyzing runtimes.}\\
\hline
\end{tabular}
\end{center}
\caption{\label{tab:Summary_example} Summaries generated by the three variants of PG-net for the same CS reflection document, and synthetic sample generated by the proposed template model. }
\end{table}
To observe the impact of using out-of-domain (news) data for pretraining to compensate for low resource in-domain (reflection) data, we train 3 variants of PG-net: model training on CNN/DM; model training on reflections; and  model training on CNN/DM then tuning using reflections. Table~\ref{tab:Summary_example} shows example summaries generated by the three variants of PG-net for a CS document. 
For all experiments where reflections are used for training/tuning, we train using a leave one course out approach (i.e, in each fold, three courses are used for training and the remaining course  for testing). If the experiment involves tuning a combined dictionary of CNN/DM and reflections is used to avoid domain mismatch. To tune model parameters, the best number of steps for training, the learning rate, etc., a randomly selected 50\% of the training data is used for validation.  We choose the parameters that maximize ROUGE scores over this validation set.

To implement PG-net we use OpenNMT \cite{2017opennmt} with the original set of parameters.
The out-of-domain model is trained for 100k steps using the CNN/DM dataset. Following base model training, we tune the model by training it using student reflections. The tuning is done by lowering the LR from 0.15 to 0.1 and training the model for additional 500 steps. The in-domain model is trained only using reflections. We use the same model architecture as above and train the model for 20k steps using adagrad and LR of 0.15. 
\\
\textbf{Synthesis Baseline (for testing H3, H4)}.
Following  \cite{zhang2015character}, we developed a data synthesis baseline using word replacement via WordNet. The baseline iterates over all words in a summary. If word $X$ has $N$ synonyms in WordNet, the model creates $N$ new versions of the summary and corresponding reflections by replacing the word $X$ with each of the $N$ synonyms.
\\
\textbf{Template Synthesis Model (for testing H4, H5)}.
To synthesize summaries, we use the same leave one course out approach. For each course, we use the data from the other three courses to train the rewriting module and tune the scoring language model. We can also use the summaries from CNN/DM data as additional samples to further train the rewriting module.
We then start synthesizing data using that training data as input. First templates are constructed. 
The templates are then clustered into 8 clusters. We decided to use 8 to avoid clustering templates from POI with MP, as the templates from both prompts would contain very different supporting words. We also wanted to avoid a high level of dissimilarity within each cluster, and allow some diversity. Following the clustering, the rewriting model produces candidate summaries for each human summary. The rewriting model is another PG-net with the same exact parameters.
After producing the candidate summaries, a language model is  used to score them. The language model is a single layer LSTM language model trained on 36K sentences from Wikipedia and fine tuned using student reflections. In this work we decided to pick only the highest 3 scored candidate summaries as synthetic data, to avoid adding ill-formed summaries to the training data. Since we are adding $N$ synthetic summaries for each set of reflections, that means we are essentially duplicating the size of our original reflection training data by $N$, which is 3 in our case.\footnote{We plan to explore the effect of varying $N$ in the future.} 
Table \ref{tab:Summary_example} shows a human summary, the keywords extracted, then the output of injecting keywords in a different template using rewriting.
\\
\textbf{Template-based Summarization (for testing H6)}.
While the proposed template-based model was intended for data synthesis, 
with minor modification it can be adapted for summarization itself. Because the modifications introduce few parameters, the model is suitable for small datasets.
Recall that for  data synthesis, the input to the template method is a summary.  Since for  summarization the input instead is a set of reflections, we perform keyword extraction over the set of reflections. We then add an extra logistic regression classifier that uses the set of reflections as input and predicts a cluster of templates constructed from other courses. Using the keywords and the predicted cluster of templates, we use the same rewriting model to produce candidate summaries. The last step in the pipeline is scoring. In data synthesis, a reference summary is used for scoring; however, in summarization we don't have such a reference. To score the candidate summaries, the model only uses the language model and produces the candidate with the highest score. 

\section{Results}
\textbf{ROUGE Evaluation Results}.
\begin{table*}[t!]
\begin{center}
\small
\begin{tabular}{|l|l|c|c|c||c|c|c||c|}
\hline
\multicolumn{2}{|c|}{\textbf{Summarization Model}} & \bf R-1 & \bf R-2 & \bf R-L & \bf R-1 & \bf R-2 & \textbf{R-L}& \textbf{Row}\\
\cline{3-9} 
& & \multicolumn{3}{|c||}{\bf CS} & \multicolumn{3}{|c||}{\bf ENGR}&\\
\hline
\multirow{4}{*}{\bf Extractive Baselines} & \cite{luo2015summarizing}  & 27.65 & 6.66 & 22.76 & \underline{30.99} & \underline{8.97} & \underline{25.38} & 1\\
\cline{2-9}
& Mead 5 & \underline{30.59} & \underline{8.26} & \underline{23.78} & 29.35 & 7.91 & 23.12 & 2\\
\cline{2-9}
& Random Select 5  & 26.74 & 5.89 & 20.55 & 26.14  & 5.35 & 20.57 & 3\\
\cline{2-9}
& Fast-RL (Extractive) & 28.95 & 6.62 & 22.16 & 26.6 & 5.09 & 21.06 & 4 \\
\hline
\hline
\multirow{7}{*}{\bf PG-net} & CNN/DM &29.83 & 7.10 & 18.28 & 29.30 & 6.95 & 17.63 & 5\\
\cline{2-9}
& Student Reflection & 25.90 & 4.62 & 17.49 & 26.14 & 6.05 & 20.94 & 6\\
\cline{2-9}
& {Student Reflection + WordNet Synthetic} & {27.15} & 3.13 & {17.8} & {28.11} & {6.11} & {21.29} & 7\\
\cline{2-9}
& {Student Reflection + Template Synthetic} & {26.93} & 3.49 & {19.38} & {29.54} & {6.96} & {21.30} & 8\\
\cline{2-9}
 & {Tuned} & \textit{37.31} & \textit{10.20}& \textit{24.16} & \textit{38.47} & \textit{13.88} & \textit{27.79} & 9\\
 \cline{2-9}
& {Tuned + WordNet Synthetic} & {34.13} & 7.13 & {21.96} & {32.61} & {7.51} & {21.72} & 10\\
 \cline{2-9}
 & {Tuned + Template Synthetic} & \underline{\textbf{\textit{37.88}}} & \underline{\textbf{\textit{11.01}}}& \underline{\textbf{\textit{25.30}}} & \underline{\textbf{\textit{38.98}}} & \underline{\textbf{\textit{13.97}}} & \underline{\textbf{\textit{28.65}}} & 11\\
\hline
\hline
\multicolumn{2}{|c|}{\textbf{Summarizing with Template Model}} & \textit{34.8} & \textit{9.3} & 23.4 & \textit{36.5} & \textit{11.2} & 24.1 & 12\\
\hline
\hline

\multicolumn{2}{|c|}{\bf Summarization Model} &  \multicolumn{3}{|c||}{\bf Stat2015} & \multicolumn{3}{|c||}{\bf Stat2016} & \\
\hline
\multirow{4}{*}{\bf Extractive Baselines} & \cite{luo2015summarizing} &  \underline{28.84} & \underline{10.15} & \underline{25.05} & \underline{32.96} & \underline{\textbf{12.44}} & \underline{27.90}  & 13\\
\cline{2-9}
& Mead 5  & 26.06 & 8.84 & 21.28 & 32.31 & 12.30 & 26.27 & 14\\
\cline{2-9}
& Random Select 5  & 23.50 & 5.88 & 19.46 & 23.77 & 7.63 & 20.11 & 15\\
\cline{2-9}
& Fast-RL (Extractive) & 27.49 & 7.73 & 22.05 & 24.59 & 8.16 & 20.66 &  16\\
\hline
\hline
\multirow{7}{*}{\bf PG-net} & CNN/DM & 27.22 & 7.62 & 17.80 & 30.99 & 10.01 & 20.29 & 17\\
\cline{2-9}
& Student Reflection & \textit{29.29} & 5.66 & 20.31 & 32.10 & 5.92 & 22.28 & 18\\
\cline{2-9}
& {Student Reflection + WordNet Synthetic} & {26.11} & 5.26 & {20.41} & {31.92} & {6.14} & {22.36} & 19\\
\cline{2-9}
& {Student Reflection + Template Synthetic} & \textit{29.65} & {5.42}& {20.54} & {32.43} & {5.96} & {21.53} & 20\\
\cline{2-9}
& {Tuned} & \textit{38.78} & \underline{\textbf{\textit{12.45}}} & \textit{26.19} & \underline{\textbf{\textit{41.05}}} & {12.17} & \textit{28.25} & 21\\
\cline{2-9}
& {Tuned + WordNet Synthetic} & {34.65} & 9.88 & {24.31} & {36.58} & {9.78} & {24.08} & 22\\
\cline{2-9}
& {Tuned + Template Synthetic} & \underline{\textbf{\textit{39.12}}} & \textit{12.23}& \underline{\textbf{\textit{26.93}}} & \textit{40.95} & \underline{12.26} & \underline{\textbf{\textit{28.51}}} & 23\\
\hline
\hline
\multicolumn{2}{|c|}{\textbf{Summarizing with Template Model}} & \textit{35.6} & \textit{11.1} & 24.8 & \textit{38.3} & 11.9 & 25.6 & 24\\
\hline
\end{tabular}
\end{center}
\caption{\label{tab:models_results} ROUGE results. \textit{Italics}  indicates outperforms 
baselines. \textbf{Boldface} indicates  best result over all models. \underline{Underlining} indicates  best result within model group (i.e., Extractive Baselines,  PG-net).}
\end{table*}
Table \ref{tab:models_results} presents  summarization performance results for the 4 extractive baselines,  for the original and proposed variants of PG-net, and finally for template-summarization. Following \cite{see2017get},  performance is  evaluated using  ROUGE (1, 2, and $L$) \cite{lin2004rouge} on F1. The motivation for using domain transfer and data synthesis is our hypothesis \textbf{(H1)}.
Table~\ref{tab:models_results} supports this hypothesis. All  ROUGE scores for PG-net that outperform all extractive baselines (in italics) involve tuning and/or use of synthesised data, except for one R-1 (row 18).

As for our second hypothesis \textbf{(H2)}, table~\ref{tab:models_results} shows that it is a valid one. For PG-net, comparing the CNN/DM out-of-domain and Student Reflection in-domain results in rows (5 and 6) and (17 and 18) with their corresponding tuned results in rows 9 and 21, we see that fine tuning  improves R-1, R-2, and R-$L$
for all courses (rows 5, 6, 9 and 17, 18, 21).  Qualitatively, the examples  presented  in Table \ref{tab:Summary_example} clearly show that  tuning yields a more coherent and relevant summary. 
Over all courses, the tuned version of PG-net  consistently outperforms the best baseline result for each metric (rows 9 vs. 1, 2, 3, 4 and 21 vs. 13, 14, 15, 16) except for R-2 in Stat2016.

To validate our next set of hypothesises \textbf{(H3, H4. H5)},
we use the synthesized data in two settings: either using it for training (rows 7, 8 and 19, 20) or tuning (rows 10, 11 and 22, 23). Table~\ref{tab:models_results} 
supports {\bf H4} by showing that the proposed synthesis model outperforms the WordNet baseline in training (rows 7, 8 and 19, 20) except Stat2016, and tuning (10, 11 and 22, 23)  over all courses.  It also shows that while adding synthetic data from the baseline is not always helpful, adding synthetic data  from the template model helps to improve both the training and the tuning process. In both CS and ENGR courses, tuning with synthetic data enhances all ROUGE scores compared to tuning with only the original data. (rows 9 and 11). As for Stat2015, R-1 and R-$L$ improved, while R-2 decreased. For Stat2016,  R-2 and R-$L$ improved, and R-1 decreased (rows 21 and 23). Training with both student reflection data and synthetic data compared to training with only student reflection data yields similar improvements, supporting \textbf{H3} (rows 6, 8 and 18, 20). While the increase in ROUGE scores is small, our results show that enriching training data with synthetic data can  benefit both the training and tuning of other models. In general, the best results are obtained when using data synthesis for both training and tuning (rows 11 and 23), supporting \textbf{H5}.

Finally, while the goal of our template model was to synthesize data, using it for summarization is surprisingly competitive, supporting \textbf{H6}. We believe that training the model with little data is doable due to the small number of parameters  (logistic regression classifier only). While rows 12 and 24 are never the best results, they are close to the best involving tuning. This encourages us to enhance our template model and explore templates not so tailored to our data.\\
\textbf{Human Evaluation Results}.
While automated evaluation metrics like ROUGE measure lexical similarity between  machine and human  summaries, humans can better measure how coherent and readable a summary is. Our evaluation study investigates whether  tuning the PG-net model increases summary coherence, by asking evaluators to select which of three summaries for the same document they like most: the PG-net model trained on CNN/DM; the model trained on student reflections; and finally the model trained on CNN/DM and tuned on student reflections.
20 evaluators were recruited from our institution and asked to each perform 20 annotations.
Summaries are presented to evaluators in random order. Evaluators are then asked to select the summary they feel to be most readable and coherent. Unlike ROUGE, which measures the coverage of a generated summary relative to a reference summary, our evaluators  don't read the reflections or reference summary. They choose the summary that is most coherent and readable, regardless of the source of the summary.
For both courses, the majority of selected summaries were produced by the tuned model (49\% for CS and 41\% for Stat2015), compared to (31\% for CS and 30.9\% for Stat2015) for CNN/DM model, and (19.7\% for CS and 28.5\% for Stat2015) for student reflections model. These results again suggest that domain transfer can remedy the size of in-domain data and improve performance.



\section{Conclusions and Future Work}
We explored improving the performance of  neural
abstractive summarizers 
when applied to the low resource domain of student reflections using three approaches: domain transfer, data synthesis and the combination of both. For domain transfer, state of the art abstractive summarization model  was pretrained using out-of-domain data (CNN/DM),  then tuned using in-domain data (student reflections). The process of tuning improved  ROUGE scores on the student reflection data, and at the same time produced more readable summaries. To incorporate synthetic data, we proposed a new template based synthesis model  
to synthesize new summaries. 
We showed that enriching the training data 
with this synthesized data can further increase the benefits of using domain transfer / tuning to increase ROUGE scores. We additionally showed that the proposed synthesis model outperformed a word replacement synthesis baseline. 

Future plans include trying domain adaptation, enhancing the synthesising process by using other models,  
further exploring template-based methods,
and extending the analysis of the synthesis model to cover other types of data like reviews and opinions.

\section{Acknowledgments}
The research reported here was supported, in whole or in part, by the institute of Education Sciences, U.S. Department of Education, through Grant R305A180477 to the University of Pittsburgh. The opinons expressed are those of the authors and do not represent the views of the institute or the U.S. Department of Education

\bibliographystyle{aaai}
\bibliography{aaai-bib}

\end{document}